\documentclass{ieeeaccess}
\usepackage{cite}
\usepackage{amsmath,amssymb,amsfonts}
\usepackage{bm}
\usepackage{braket}
\usepackage{algorithmic}
\usepackage{graphicx}
\usepackage{dblfloatfix}
\usepackage{textcomp}
\usepackage{booktabs}
\usepackage{framed}

\def\BibTeX{{\rm B\kern-.05em{\sc i\kern-.025em b}\kern-.08em
    T\kern-.1667em\lower.7ex\hbox{E}\kern-.125emX}}
\graphicspath{{../fig}}

\begin{document}
\history{Date of publication xxxx 00, 0000, date of current version xxxx 00, 0000.}
\doi{xx.xxxx/ACCESS.20xx.DOI}

\onecolumn
\begin{framed}
    \noindent
    \copyright 2022 IEEE. Personal use of this material is permitted. Permission from IEEE must be obtained for all other uses, in any current or future media, including reprinting/republishing this material for advertising or promotional purposes, creating new collective works, for resale or redistribution to servers or lists, or reuse of any copyrighted component of this work in other works.%
\end{framed}
\clearpage
\twocolumn

\title{Black-box optimization for integer-variable problems using Ising machines and factorization machines}
\author{%
    Yuya Seki\authorrefmark{1},
    Ryo Tamura\authorrefmark{2,3,4},
    and Shu Tanaka\authorrefmark{5,6,7,8}
}
\address[1]{%
Graduate School of Science and Technology, Keio University, Kanagawa 223-8522, Japan (e-mail: yuya.seki@keio.jp)}
\address[2]{%
International Center for Materials Nanoarchitectonics(WPI-MANA),
National Institute for Materials Science, Ibaraki 305-0044, Japan (e-mail: tamura.ryo@nims.go.jp)}
\address[3]{%
Research and Services Division of Materials Data and Integrated System, National Institute for Materials Science 1-1 Namiki, Tsukuba-city, Ibaraki, 305-0044, Japan}
\address[4]{%
Graduate School of Frontier Sciences, The University of Tokyo, Chiba 277-8568, Japan}
\address[5]{%
Department of Applied Physics and Physico-Informatics, Keio University, Kanagawa 223-8522, Japan
(e-mail: shu.tanaka@appi.keio.ac.jp)}
\address[6]{%
Quantum Computing Center, Keio University, Kanagawa 223-8522, Japan}
\address[7]{Green Computing System Research Organization, Waseda University, Tokyo, 162-0042, Japan}
\address[8]{%
Quantum Computing Unit, Institute of Innovative Research, Tokyo Institute of Technology, Tokyo 105-0023, Japan}
\tfootnote{
    This work was supported in part by the Cross-ministerial Strategic Innovation Promotion Program (SIP)
    (Photonics and Quantum Technology for Society 5.0).
}

\markboth
{Yuya Seki \headeretal: Black-box optimization for integer-variable problems using Ising machines and factorization machines}
{Yuya Seki \headeretal: Black-box optimization for integer-variable problems using Ising machines and factorization machines}

\corresp{Corresponding author: Yuya Seki (e-mail: yuya.seki@keio.jp).}

\begin{abstract}
Black-box optimization has potential in numerous applications such as hyperparameter optimization in machine learning and optimization in design of experiments.
Ising machines are useful for binary optimization problems because variables can be represented
by a single binary variable of Ising machines.
However, conventional approaches using an Ising machine cannot handle black-box optimization problems with non-binary values.
To overcome this limitation, we propose an approach for integer-variable black-box optimization problems by using Ising/annealing machines and factorization machines in cooperation with three different integer-encoding methods.
The performance of our approach is numerically evaluated with different encoding methods
using a simple problem of calculating the energy of the hydrogen molecule in the most stable state.
The proposed approach can calculate the energy using any of the integer-encoding methods.
However, one-hot encoding is useful for problems with a small size.
\end{abstract}

\begin{keywords}
% Enter key words or phrases in alphabetical 
% order, separated by commas. For a list of suggested keywords, send a blank 
% e-mail to keywords@ieee.org or visit \underline
% {http://www.ieee.org/organizations/pubs/ani\_prod/keywrd98.txt}
Encoding,
Machine learning,
Optimization,
Physics
\end{keywords}

\titlepgskip=-15pt

\maketitle

\section{Introduction}
\label{sec:introduction}
\PARstart{M}{any} recent studies have investigated Ising and annealing machines~\cite{Johnson2011quantum,Barends2016digitized,inagaki2016coherent,Rosenberg20173dintegrated,Maezawa2019toward,Novikov2018exploring,Mukai2019superconducting,Tsukamoto2017accelerator,Aramon2019,Yamaoka2016,Goto2019combinatorial,mohseni2022ising,Oku2022how}.
These machines are utilized to search for low energy solutions of Ising problems and quadratic unconstrained binary optimization (QUBO) problems~\cite{Lucas2014,tanaka2017quantum,tanahashi2019application}.
These efforts have expanded the scope and application of Ising machines to real-world problems
because Ising and QUBO problems cover a wide range of combinatorial optimization problems~\cite{neukart2017traffic,Terada2018an,nishimura2019item,ohzeki2019control,nath2021review,Bao2021multiday,mukasa2021ising}.

A drawback of Ising machines is that they cannot represent combinatorial optimization problems with cost functions of binary variables.
To apply Ising machines, the cost function of binary variables must be manually constructed from the problem settings of the combinatorial optimization problem.
This process can be difficult.
It is impossible to construct cost functions for black-box optimization problems since the analytical form of the black-box function is unknown.
Black-box optimization problems include important problems such as the optimization of hyperparameters in deep neural networks~\cite{Eggensperger2013towards} and automated material discovery based on existing data about the properties of materials~\cite{Ju2017,Butler2018,Terayama2021blackbox}.
Conventional approaches using Ising machines cannot handle such black-box optimization problems.

A Factorization Machine with Quantum Annealing (FMQA) is a prospective approach for black-box optimization problems using Ising machines~\cite{Kitai2020designing}.
FMQA is classified as a type of surrogate modeling, in which a black-box function is approximated by a tractable model called a surrogate model~\cite{Brochu2010tutorial,Audet2000surrogate}.
By obtaining low-cost solutions of the surrogate model instead of the black-box function, the black-box optimization problem can be solved.
However, finding optimal solutions in the surrogate model is a difficult task when the black-box optimization treats a vast exploration space.
To allow Ising machines to handle this optimization task, FMQA uses a Factorization Machine (FM)~\cite{Rendle2010factorization} with binary variables as a surrogate model.
Since the model equation of FM with binary variables takes the QUBO form, Ising machines can be employed to obtain low-cost solutions of the surrogate model.
Kitai \textit{et al.}\ reported that FMQA discovered a radiator with a wavelength selectivity higher than previously designed ones~\cite{Kitai2020designing}.
In subsequent studies, FMQA evolved into a black-box optimization method based on a quantum circuit model~\cite{Gao2021quantumclassical}.
Additionally, other methods for black-box optimization problems with binary variables using Ising machines have been developed~\cite{koshikawa2021,sato2021}.

Herein we expand the application range of FMQA.
Previous studies dealt with problems in which each variable of the black-box function takes two values.
In this situation, a variable can be represented by a single binary variable of Ising machines. However, a more general black-box function accepts variables with values more than two.
Hereafter, general problems are referred to as integer-variable black-box optimization problems.
Typically, integer variables are represented by encoding methods with binary variables (e.g., binary encoding, one-hot encoding, and domain-wall encoding), where multiple binary variables of an Ising machine are assigned to a single integer variable~\cite{rosenberg2016solving,tanahashi2019application,tamura2021performance,zaman2022pyqubo}.

\begin{figure}[t]
    \centering
    \includegraphics[width=3.2truein]{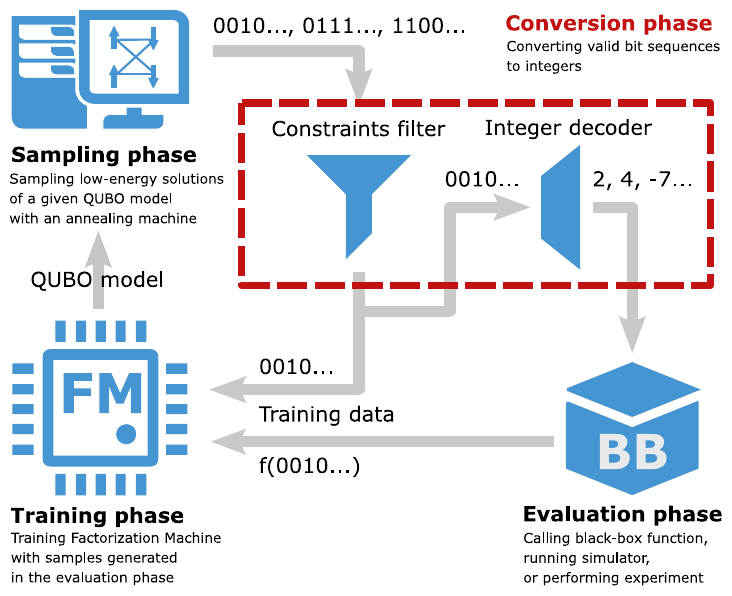}
    \caption{%
        Schematic picture of our method.
        Procedure consists of four phases:
        the evaluation phase, training phase, sampling phase,
        and conversion phase.
        Gray arrows represent the flow of data.
    }
    \label{fig:scheme}
\end{figure}

This paper reveals the properties of different encoding methods in the context of FMQA.
Schematic picture of our method is shown in Fig.~\ref{fig:scheme}.
Encoding methods are integrated into FMQA, and their performances are evaluated for a simple problem.
In Ref.~\cite{Kitai2020designing}, quantum annealing (QA) is used to obtain low-cost solutions of the surrogate model.
By contrast, the conventional simulated annealing (SA) method is used instead of QA because SA is unaffected by the noise of actual hardware devices.
To exclude unknown effects of noise and to clarify the properties of integer-encoding methods, herein we adopt SA, whose nature is studied well.
We refer to the method where the annealer in FMQA is not restricted to QA as Factorization Machine with Annealing (FMA).
It should be noted that Ising machines can be applied to continuous black-box optimization problems using the encoding method developed in Ref.~\cite{izawa2022continuous}.
Thus, combining our method with the above one, it is possible to apply FMA to both discrete and continuous black-box optimization problems.

The rest of this paper is organized as follows.
Section~\ref{sec:fm} provides a brief review of FM.
Section~\ref{sec:method} proposes our optimization approach for integer-variable black-box optimization problems based on FMA.
Section~\ref{sec:encodings} describes the encoding methods used in this paper.
Section~\ref{sec:application} details the numerical experiments to investigate the properties of the proposed approach, and the results are shown in Sec.~\ref{sec:results}.
Finally, Sec.~\ref{sec:conclusion} concludes this paper.

\section{FM\MakeLowercase{s}}
\label{sec:fm}
FMs are machine learning models that combine a support vector machine and a factorization model~\cite{Rendle2010factorization}.
Unlike a support vector machine, FM can achieve a high performance even for sparse data due to the dependent interaction coefficients in the model equation.

The model equation of FM for two degrees is defined as
\begin{align}
    y(\boldsymbol{x}; \boldsymbol{\theta}) = c + \sum_{i=1}^{N}q_{i}x_{i}
    + \sum_{\substack{i,j=1 \\ (i < j)}}^{N}
    \left\langle \boldsymbol{v}_{i}, \boldsymbol{v}_{j} \right\rangle x_{i}x_{j}.
    \label{eq:fm-model-eq}
\end{align}
Here, $\boldsymbol{x} \equiv (x_{1}, \dotsc, x_{N})$ denotes the explanatory variables.
This paper considers only FM with binary variables, $\boldsymbol{x}\in \{0, 1\}^{N}$.
The parameters $c\in \mathbb{R}$, $q_i \in \mathbb{R}\ (i=1, \dotsc, N)$, and $\boldsymbol{v}_{i} \in \mathbb{R}^{k}\ (i=1, \dotsc , N)$
are model parameters to be trained.
The list of all the model parameters is denoted by $\boldsymbol{\theta}$.
The symbol $\langle \cdot, \cdot \rangle$ denotes the inner product.
The integer $k$ controls the expression ability of the model.
The inner product $\langle \boldsymbol{v}_{i}, \boldsymbol{v}_{j} \rangle$ can express any second-order coefficient matrix in the limit of large $k$.
However, when less training data is available, FM with a small $k$ performs well because overfitting is suppressed.
Hereafter, the parameter $k$ is referred to as the rank of FM.
The summation of the second-order term is taken over all the possible pairs of $(i,j)$
that satisfy $1 \le i < j \le N$.

The model equation with binary variables takes the QUBO form.
Let $Q = (Q_{i,j})$ be an $N \times N$ matrix defined as
\begin{align}
    Q_{i,j} \equiv \begin{cases}
        q_{i} & (i=j), \\
        \left\langle \boldsymbol{v}_{i}, \boldsymbol{v}_{j} \right\rangle & \text{otherwise}.
    \end{cases}
\end{align}
The model equation in Eq.~\eqref{eq:fm-model-eq} can be rewritten using $Q$ as
\begin{align}
    y(\boldsymbol{x}; \boldsymbol{\theta}) = c
    + \sum_{\substack{i,j=1 \\ (i \le j)}}^{N} Q_{i,j} x_{i}x_{j}.
    \label{eq:fm-qubo}
\end{align}
Here, the diagonal elements correspond to the linear term in Eq.~\eqref{eq:fm-model-eq}
since $Q_{i,i} (x_{i})^{2} = q_{i}x_{i}$.
Therefore, the model equation with binary variables takes the QUBO form in Eq.~\eqref{eq:fm-qubo}.

\section{Proposed approach}
\label{sec:method}
Here, we propose an optimization approach for integer-variable black-box optimization problems based on FMA.
Fig.~\ref{fig:scheme} shows the flow diagram of our approach.
The proposed approach is composed of four phases: the evaluation phase, training phase, sampling phase, and conversion phase.
By iterating the sequence of the four phases, the target black-box function is approximated as a QUBO model.
The optimal solution of the target black-box function is determined by finding the low-energy state of the QUBO model.

In the training phase, FM is trained using the obtained samples.
In the first iteration, arbitrary samples of a pair of bit sequence expressing the integer explanatory variables and corresponding cost must be prepared.
An encoding method, which is explained in the next section, is used to convert integers into binary variables.
A model optimizer finds the parameters in FM to minimize the mean-squared error between the values of the model equation and the costs of the samples.
Common optimizers include the momentum method~\cite{Rumelhart1986} and the Adams method~\cite{Kingma2014adam}.
Beginning in the second iteration, the model is updated using all the samples obtained at that point.

In the sampling phase, new samples are generated from the trained FM.
To extract information of the low-cost regime of the black-box function, samples are generated with low values of the model equation of FM.
Since the model equation of FM is exactly in the QUBO form, an annealer can generate low-value states of the trained FM model. 

In the conversion phase, samples of bit sequences are converted into integers.
First, infeasible samples are removed.
An annealing machine such as SA can output invalid samples.
For instance, in the cases of one-hot encoding or domain-wall encoding, samples may violate the constraints for the encoding methods.
Additionally, even if the samples satisfy the constraints, they may not be suitable for the black-box function.
Such a situation occurs when calculating the ground-state energy of hydrogen molecule (Sec.~\ref{sec:blackbox-h2}).
Then, valid samples are converted into integers via an integer decoder (Sec.~\ref{sec:encodings}).

In the evaluation phase, a black-box function is invoked to evaluate the actual costs for the samples generated in the sampling phase.
The obtained pairs of the samples and corresponding values of the cost function are appended to the sample set.
The above sequence is iterated a certain number of times.
After the end of the iteration, the best sample in the set is outputted.

\section{Integer-encoding methods}
\label{sec:encodings}
Here, the conversion of a bit sequence into an integer is considered using three different encoding methods: binary encoding, one-hot encoding, and domain-wall encoding.
Tables~\ref{table:binary-encoding}--\ref{table:domainwall-encoding} show examples of integer encoding from $-2$ through $1$ using the different methods.
In the following, the equations that convert a given bit sequence into an integer are explained and the cost functions in the QUBO form, which are passed to an annealing machine, are shown for each encoding method.

\begin{table}
\setlength{\tabcolsep}{3pt}
\begin{minipage}{0.31\linewidth}
    \centering
    \caption{\\Binary encoding}
    \label{table:binary-encoding}
    \begin{tabular}{cc}
    \toprule
    Integer& 
    Bit sequence \\
    \midrule
    $-2$& $10$ \\
    $-1$& $11$ \\
    $0$& $00$ \\
    $1$& $01$ \\
    \bottomrule
    \end{tabular}
\end{minipage}
\hfill
\begin{minipage}{0.31\linewidth}
    \centering
    \caption{\\One-hot encoding}
    \label{table:onehot-encoding}
    \begin{tabular}{cc}
    \toprule
    Integer& 
    Bit sequence \\
    \midrule
    $-2$& $1000$ \\
    $-1$& $0100$ \\
    $0$& $0010$ \\
    $1$& $0001$ \\
    \bottomrule
    \end{tabular}
\end{minipage}
\hfill
\begin{minipage}{0.31\linewidth}
    \centering
    \caption{\\Domain-wall encoding}
    \label{table:domainwall-encoding}
    \begin{tabular}{cc}
    \toprule
    Integer& 
    Bit sequence \\
    \midrule
    $-2$& $000$ \\
    $-1$& $100$ \\
    $0$& $110$ \\
    $1$& $111$ \\
    \bottomrule
    \end{tabular}
\end{minipage}
\end{table}

\subsection{Binary encoding}
Binary encoding is a basic encoding method to represent integers.
A signed integer $n$ can be expressed by a bit sequence $\bm{x} = (x_0, \dotsc ,x_{d-1})$,
where $d$ is the length of the bit sequence, as
\begin{align}
    n = x_0 + 2 x_1 + \dotsb + 2^{d-2} x_{d-2} - 2^{d-1}x_{d-1}.
    \label{eq:binary_encoding}
\end{align}
The variable $n$ can inclusively take any integer from $-2^{d-1}$ through $2^{d-1}-1$.
Here, consider a QUBO model for a black-box function with $L$ integer-valued variables,
$n^{(1)}, \dotsc , n^{(L)}$.
The Hamiltonian of the QUBO model based on binary encoding is given as
\begin{align}
    H = \sum_{l, m = 1}^{L}\sum_{\substack{i,j=0 \\ (i\le j)}}^{d-1}
        Q_{i,j}^{l,m}x_{i}^{(l)}x_{j}^{(m)} \equiv H_{\text{FM}},
    \label{eq:binary_QUBO}
\end{align}
where the variable $x_{i}^{(l)}$ denotes the $i$-th bit to represent the $l$-th variable $n^{(l)}$.
The summation for $i$ and $j$ is taken over all the possible pairs of $(i, j)$ satisfying
$0\le i \le j \le d-1$.
Throughout this paper, the constant term of the Hamiltonian is omitted because it is irrelevant for our proposed approach.

The Hamiltonian $H_{\text{FM}}$ is constructed by training FM.
The model parameters for the quadratic term in FM and the model parameters for the linear term correspond to the off-diagonal elements of $Q = (Q_{i,j}^{l,m})$ and the diagonal elements, respectively.
To normalize the energy scale of the model in Eq.~\eqref{eq:binary_QUBO}, the original matrix $Q$ is multiplied by a normalization factor $1 / \max \lvert Q_{i,j}^{l,m}\rvert$.
Hereafter, the normalized matrix $Q$ is used for the cost function.

\subsection{One-hot encoding}
One-hot encoding is another famous method.
In this method, an integer is represented by the position of only an active bit.
An integer is represented by a bit sequence as
\begin{align}
    n = n_0 + \sum_{i=0}^{d-1} i x_{i},
    \label{eq:one-hot_encoding}
\end{align}
under a constraint such that
\begin{align}
    \sum_{i=0}^{d-1} x_{i} = 1.
    \label{eq:one-hot_constraint}
\end{align}
The parameter $n_{0}$ in Eq.~\eqref{eq:one-hot_encoding} is an integer,
which shifts the range of integers represented by one-hot encoding.
The variable $n$ in Eq.~\eqref{eq:one-hot_encoding} can take any integer from $n_{0}$ through $n_{0} + d - 1$.
The bit sequence with $d$ bits can represent $d$ integers,
which is much smaller than binary encoding.
In the binary-encoding method, a $d$-bit sequence can represent $2^{d}$ integers.
Although this may appear to be a drawback, one-hot encoding has an advantage in the context of FMA, as shown later.

The QUBO model in the one-hot FMA consists of two terms: one for FM and another for the one-hot constraint term given by Eq.~\eqref{eq:one-hot_constraint}.
The model is given as
\begin{align}
    H = H_{\text{FM}} + p C_{\text{oh}}\left(\{\bm{x}^{(l)}\}_{l=1}^{L}\right),
    \label{eq:one-hot_QUBO}
\end{align}
where $H_{\text{FM}}$ is the same as that in Eq.~\eqref{eq:binary_QUBO}, except
that the variables $x_{i}^{(l)}$ represent binary variables for an $l$-th integer variable
in the one-hot encoding given by Eq.~\eqref{eq:one-hot_encoding}.
The constraint term is necessary to avoid generating infeasible samples.
The parameter $p$ in Eq.~\eqref{eq:one-hot_QUBO} is a positive real number,
which controls the strength of the one-hot constraint,
and the constraint term is given as
\begin{align}
    C_{\text{oh}}\left(\{\bm{x}^{(l)}\}_{l=1}^{L}\right) = \sum_{l=1}^{L}\left(
        \sum_{i=0}^{d-1}x_{i}^{(l)} - 1
    \right)^{2}.
    \label{eq:one-hot_constrait_QUBO}
\end{align}
Due to the constraint term, costs for infeasible samples increase by at least $p$.

\subsection{Domain-wall encoding}
Domain-wall encoding represents integers by bits of sequential active bits
followed by inactive bits~\cite{chancellor2019domain}.
The position of the boundary between the active bits and inactive bits,
which is called the domain wall, corresponds to an integer.
In addition to the bit sequences with a domain wall, bit sequences with all active bits
and all inactive bits are used to represent the largest integer and smallest integers, respectively.
A bit sequence with one domain wall, at most, represents an integer as
\begin{align}
    n = n_0 + \sum_{i=1}^{d-1}i(x_{i-1} -2 x_{i-1} x_i + x_i) + dx_{d-1},
    \label{eq:domain-wall_enconding}
\end{align}
where $n_0$ is the shift parameter for the representable range of integers.
The variable $n$ in Eq.~\eqref{eq:domain-wall_enconding} can take any integer between
$n_0$ and $n_0 + d$.

The QUBO model for domain-wall FMA is given as
\begin{align}
    H = H_{\text{FM}} + p C_{\text{dw}}\left(\{\bm{x}^{(l)}\}_{l=1}^{L}\right),
    \label{eq:domain-wall_QUBO}
\end{align}
with
\begin{align}
    C_{\text{dw}}\left(\{\bm{x}^{(l)}\}_{l=1}^{L}\right)
    =
    2\sum_{l=1}^{L}\left(
        \sum_{i=1}^{d-1}x_{i}^{(l)} - \sum_{i=0}^{d-2}x_{i}^{(l)}x_{i+1}^{(l)}
    \right).
    \label{eq:domain-wall_constraint_QUBO}
\end{align}
The term $H_{\text{FM}}$ in Eq.~\eqref{eq:domain-wall_QUBO} is the same as that in Eq.~\eqref{eq:binary_QUBO},
except that the variables $x_{i}^{(l)}$ represent a binary variable for an $l$-th integer variable in the domain-wall encoding in Eq.~\eqref{eq:domain-wall_enconding}.
The constraint term in Eq.~\eqref{eq:domain-wall_constraint_QUBO} adds a cost of unity
per an extra domain wall.
Since the number of domain walls only changes by even numbers, the smallest cost
caused by the constraint in Eq.~\eqref{eq:domain-wall_QUBO} is $2p$.

\section{Setup of numerical experiments}
\label{sec:application}
To elucidate the properties of the integer-encoding methods in the context of FMA, a benchmark problem is constructed.
The benchmark calculates the ground-state energy of the hydrogen molecule.
This problem is adopted for the following reasons.

First, an explicit Hamiltonian of the hydrogen molecule can be derived.
Hence, a function that receives a quantum state can readily be defined,
and an expectation value of the Hamiltonian can be returned with regard to the input quantum state. 
Strictly speaking, the function is not a black-box function.
Herein it is used as a black-box function because it is possible to evaluate the energy error of FMA using an explicit Hamiltonian.
Section~\ref{sec:blackbox-h2} describes the Hamiltonian of the hydrogen molecule and the black-box function.

Second, the search-space dimension for the black-box function can be reduced using knowledge about the hydrogen molecule.
Optimizing the black-box function with a full search space should be more difficult than that with a reduced search space.
The dependency of FMA on the search-space dimension can be checked consistently.
Hence, this paper considers the dimensions of $2$ and $6$.
Section~\ref{sec:reducltion-dimension} describes how to reduce the search space.
Finally, the last part of this section details the setup of parameters in FMA.

\subsection{Black-box function}
\label{sec:blackbox-h2}
The second quantized Hamiltonian of the hydrogen molecule is given by
\begin{align}
    H = h_0 + \sum_{p,q=1}^{4}h_{pq}c_{p}^{\dagger}c_{q}
    + \sum_{p,q,r,s=1}^{4}h_{pqrs}c_{p}^{\dagger}c_{q}^{\dagger}c_{r}c_{s},
    \label{eq:hydrogen-hamiltonian}
\end{align}
where $h_{0}$ is a constant stemming from the potential energy between the atomic nuclei of hydrogen.
The overlap integrals $h_{pq}$ and $h_{pqrs}$ are readily calculated by classical computers~\cite{Reeves1965,Oohata1966gaussian}.
The operator $c_{p}^{\dagger}$ ($c_{p}$) denotes the fermion creation (annihilation) operator for mode $p$.
The system in Eq.~\eqref{eq:hydrogen-hamiltonian} has four fermion modes.
Hence, the quantum state can be represented as a sixteen-dimensional vector as
\begin{align}
    \ket{\Phi} = \begin{pmatrix}
        \phi_1 & \cdots & \phi_{16}
    \end{pmatrix}^{\top}
    \in
    \mathbb{R}^{16}.
    \label{eq:quantum-vector-16}
\end{align}

The procedure of the black-box function is as follows.
The function receives the elements of the vector in Eq.~\eqref{eq:quantum-vector-16}.
The overlap integrals in Eq.~\eqref{eq:hydrogen-hamiltonian} are calculated and the Hamiltonian is obtained.
Note that the values of the overlap integrals should be kept for subsequent invocations
of the black-box function to reduce the computation time.
The output of the function is the expectation value of the Hamiltonian with regards to the input quantum state ${\braket{\Phi | H | \Phi}}/{\braket{\Phi | \Phi}}$.
Since the vector in Eq.~\eqref{eq:quantum-vector-16} must be a valid quantum state,
the zero vector must be removed using a filter before the vector is passed to the black-box function.

The proposed method searches for the ground state of the Hamiltonian given by Eq.~\eqref{eq:hydrogen-hamiltonian} within the space where each element of a quantum state vector takes an integer in a limited range.
The range is controlled by the number of digits $d$.
By increasing $d$, the true ground state can be approximated with a high accuracy.

\subsection{Reduction of the search-space dimension}
\label{sec:reducltion-dimension}
Knowledge about the hydrogen molecule described by the Hamiltonian in Eq.~\eqref{eq:hydrogen-hamiltonian} can reduce the search-space dimensions.
Below, we show that some elements in the quantum state vector are zero in the ground state, reducing the dimensions to the number of nonzero elements.

Considering the number of electrons in a hydrogen molecule, the dimensions can be reduced from sixteen to six.
Because a hydrogen molecule has two electrons, it is sufficient to consider the space where the quantum state consists of two fermions.
Therefore, the dimension is reduced to $\binom{4}{2}=6$.
Moreover, the dimension can be reduced to two using the fact that only two elements of the quantum state vector are nonzero in the ground state.

\subsection{Setup of parameters in FMA}
Throughout this paper, we use initial samples in which quantum states are represented by a canonical basis, where one element is unity and the others are zero.
The number of initial samples is two for the problem with a two-dimensional search space and six for a six-dimensional search space.

The training phase adopts the Adam method with a learning rate of $0.01$.
The model parameters are updated using all the samples obtained at the time until the training error reaches $10^{-8}$ or the number of updates is $2,000$.

In the sampling phase, the SA method with a heat-bath update is used.
A suitable annealing schedule of SA is important to obtain low-cost solutions of QUBO problems.
A proper initial inverse temperature can be obtained by considering coefficients in the QUBO model (see Appendix~\ref{sec:annealing-schedule} for details).
The final inverse temperature is set to $100$, which was chosen through preliminary simulations.
In our scheduling, the inverse temperature increases geometrically from the initial value to the final value in $100$ steps.
In each step, Monte Carlo updates are performed $100$ times.
All the variables are updated with the heat-bath method in a random order for a single Monte Carlo update.
SA generates $60$ samples.
After removing the duplicates, three lower-cost samples are added to the sample set if the sample has not been generated in the procedure.

The four-phase sequence of FMA is reiterated until the number of iterations reaches $1,000$,
the number of samples in the set exceeds $1,000$, or no new samples are obtained
through SA in six consecutive iterations.
The lowest-cost sample among the set is outputted.

\section{Results}
\label{sec:results}
First, the results for the problem with a two-dimensional search space are shown.
Our approach using any of the three encoding methods can calculate the ground-state energy of the hydrogen molecule with a high accuracy for various bond lengths.
One-hot FMA can reduce the energy error effectively compared to the other FMAs.
Additionally, the behavior of FMA changes as the search-space dimensions change.

\subsection{Calculation of the ground-state energy of a hydrogen molecule represented by two variables}

\subsubsection{Ground-state energy by our proposed approach}
\label{sec:result-n2-gs-energy}
\begin{figure*}[!tb]
    \centering
    \includegraphics[width=6.8truein]{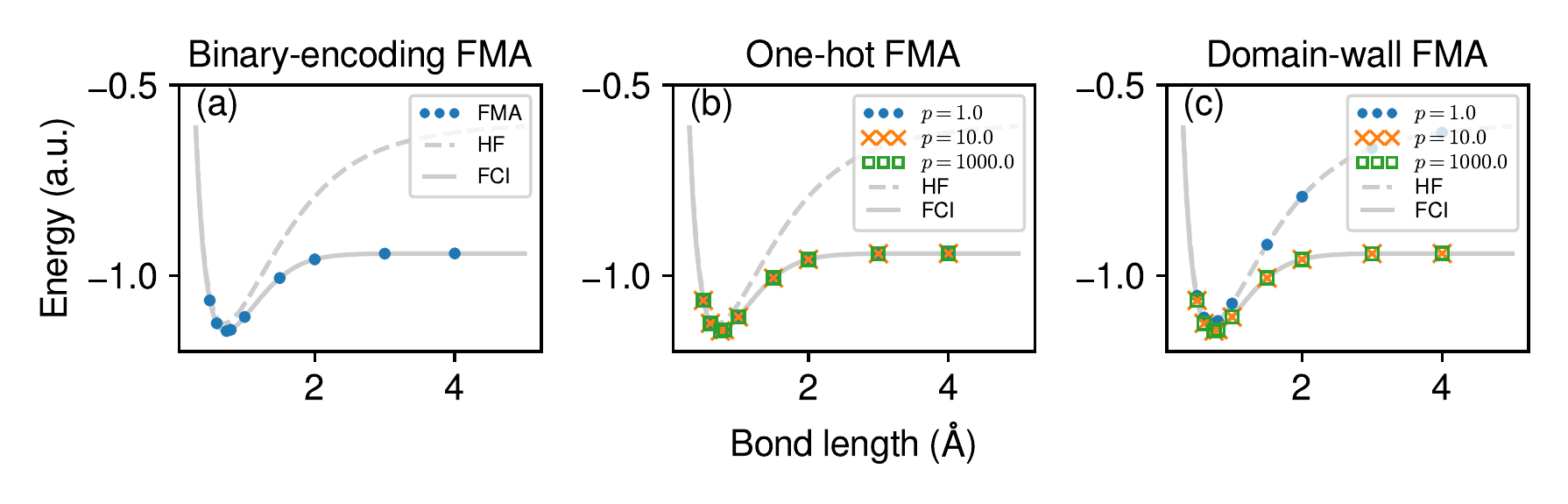}
    \caption{
        Ground-state energy of the hydrogen molecule in atomic units as a function of the bond length.
        (a) Results of FMA with binary encoding (blue dots).
        Dashed and solid gray lines represent the comparative data calculated
        by the Hartree--Fock (HF) approximation and fully configuration integration (FCI), respectively.
        (b) Results of FMA with one-hot encoding with $p=1$ (blue dots), $p=10$ (orange crosses), and $p=1,000$ (green open squares).
        (c) Results of FMA with domain-wall encoding with $p=1$ (blue dots), $p=10$ (orange crosses), and $p=1,000$ (green open squares).
    }
    \label{fig:energy-vs-bond_len}
\end{figure*}

Fig.~\ref{fig:energy-vs-bond_len} shows that the proposed approach can produce a low energy, which is close to the true ground-state energy of a hydrogen molecule calculated by fully configuration integration (FCI).
For comparison, the energies calculated by the Hartree–Fock (HF) approximation are shown.
The HF approximation gives a higher energy than the true ground-state one because the approximation does not consider the correlation between electrons.
By contrast, the FCI method takes the correlation into account.
Hence, the true ground-state energy is obtained.
The energy given by the FCI method is identical to the lowest eigenvalue of the Hamiltonian for a hydrogen molecule.
Therefore, we define the energy error of FMA as the energy difference between FMA and FCI.
To calculate an accurate energy for a given bond length, $10$ runs of FMA were performed by changing the initial condition.
That is, the initial values of the model parameters of FM and the initial states in SA were varied, and the lowest energy was taken.
Note that the HF solution is included in the initial samples.
Hence, the results of the proposed FMAs must be better than the HF solution.
Below the results for the proposed approach with different encoding methods are described in detail.

The energies calculated by the binary-encoding FMA agree well with the true ground-state energies (Fig.~\ref{fig:energy-vs-bond_len}(a)).
Here, the signed binary expression is used with $8$ bits to represent integers.
Thus, the number of variables in the QUBO model is $16$.
The $8$-bit integer takes values from $-128$ to $127$.
We construct the FM with a rank of $4$, and the number of model parameters is $80$.
The resultant energies are close to the true ground-state energy calculated by FCI in the displayed energy scale.
Sections \ref{sec:result-energy-error-dist-n2} and \ref{sec:result-blackbox-calls-n2} show a more detailed analysis of the energy error.

Second, Fig.~\ref{fig:energy-vs-bond_len}(b) shows the resultant energies obtained by FMA with one-hot encoding.
Unlike the case of binary-encoding FMA, one-hot FMA has constraints.
The parameter $p$ represents a coefficient, which controls the energy loss when the one-hot constraint is violated.
One-hot FMA is performed with penalty coefficients of $p = 1$, $10$, and $1,000$ to investigate the appropriate penalty.
An integer is represented by $64$ bits.
That is, the number of variables in the QUBO model is $128$.
The $64$-bit integer takes values from $-32$ to $31$.
The rank of FM is set to $8$.
Hence, there are $1,152$ model parameters.
The parameters for training are the same as those used for binary-encoding FMA.
Note that samples that do not satisfy the one-hot constraint are removed from the sample set.
Then the method is applied with each value of the penalty coefficient to calculate the ground-state energy of the hydrogen molecule.
As a result, the energies calculated by one-hot FMA agree well with the true ground-state energies for all the penalty coefficients studied.

Finally, Fig.~\ref{fig:energy-vs-bond_len}(c) shows the results for FMA with domain-wall encoding.
The parameter $p$ denotes the penalty coefficient similar to the case of one-hot FMA.
The number of bits to represent an integer is $63$ to cover the same range of integers as the one-hot FMA case.
The rank of FM is set to $8$.
Thus, the number of model parameters is $1,134$.
The parameters for training are the same as those for binary-encoding FMA.
Unlike one-hot FMA, domain-wall FMA does not produce energies close to the true ground-state energies when $p = 1$ because the penalty coefficient is too small to obtain samples that satisfy the domain-wall constraints.
A feasible solution is not found by SA in this case.
Consequently, the initial samples provide the lowest-cost one.
This result is identical to the HF solution.
However, the resultant energies for $p = 10$ and $p = 1,000$ agree well with the true ground-state energies.
The subsequent sections use $p = 1,000$ for domain-wall FMA.

\subsubsection{Energy error distribution}
\label{sec:result-energy-error-dist-n2}
\begin{figure*}[!tb]
    \centering
    \includegraphics[width=6.8truein]{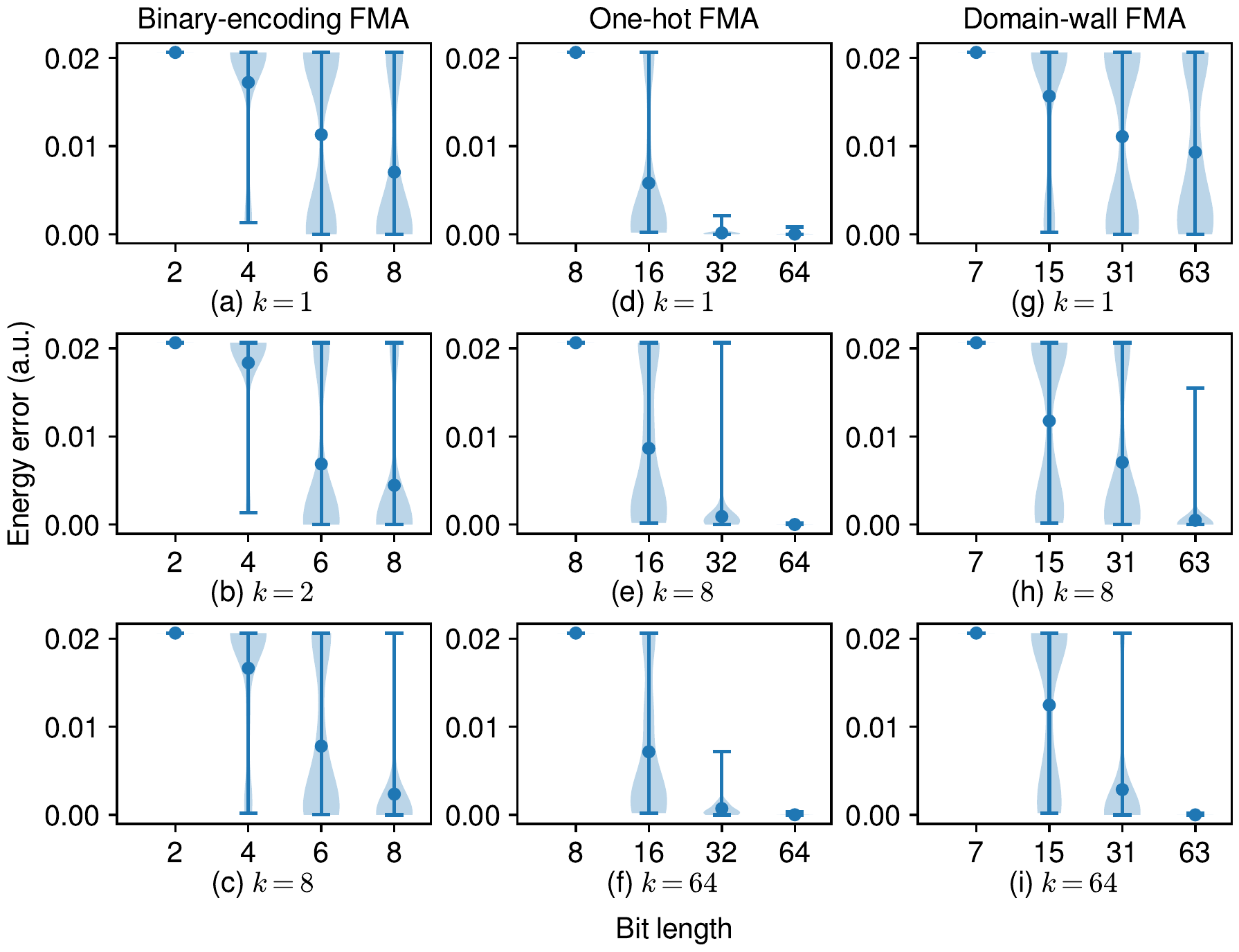}
    \caption{
        Distribution of the energy error in atomic units using our proposed FMAs.
        Width of the light blue area represents the distribution density of the energy error,
        Blue dot indicates the mean of the distribution.
        Top and bottom bars denote the largest and smallest error in the distribution, respectively.
        (a)--(c) Energy error distributions for binary-encoding FMA with bit lengths of $2$, $4$, $6$, and $8$.
        Ranks of FM are given as (a) $k=1$, (b) $k=2$, and (c) $k=8$.
        (d)--(f) Energy error distributions for one-hot FMA with bit lengths of $8$, $16$, $32$, and $64$.
        Ranks of FM are given as (d) $k=1$, (e) $k=8$, and (f) $k=64$.
        (g)--(i) Energy error distributions for domain-wall FMA with bit lengths of $7$, $15$, $31$, and $63$.
        Ranks of FM are given as (g) $k=1$, (h) $k=8$, and (i) $k=64$.
    }
    \label{fig:H2_error_dist_n2}
\end{figure*}

Fig.~\ref{fig:H2_error_dist_n2} shows the energy error distribution at the optimal bond length of a hydrogen molecule ($0.7414$ \AA) for each method as functions of the number of bits and the value of k in FM.
Here, the energy error is defined as the difference between the energy calculated by FMA and the true ground-state energy calculated by FCI.
The energy error varies by run because the result depends on the initial condition of FMA.
That is, the initial values of the model parameters of FM and the initial state of the simulated annealing method influence the result.
To obtain the energy error distribution, FMA is run $100$ times while changing the initial condition.
The procedure is reiterated $1,000$ times.
Hence, the number of samples in the set exceeds $1,000$ or until no new samples are obtained using SA for six consecutive iterations.
Similar to the calculation of the ground-state energies in Sec.~\ref{sec:result-n2-gs-energy}, the Adam method is used with a learning rate of $0.01$ and the model parameters are updated using all the samples obtained at the time until the training error reaches $10^{-8}$ or the number of updates is $2,000$.
The simulated annealing has the same setup as Sec.~\ref{sec:result-n2-gs-energy}.
From the energy calculations for the penalty coefficients in Sec.~\ref{sec:result-n2-gs-energy}, $p = 1,000$ is adopted as the penalty coefficients both for one-hot FMA and domain-wall FMA.

One-hot FMA has an advantage over the other methods.
Binary-encoding FMA tends to lower the energy error for a large bit length and a large $k$.
However, the energy error distribution is not localized in the low energy error regime.
By contrast, the energy error distribution of one-hot FMA is localized in the low energy error regime for a large bit length.
Although the domain-wall FMA with bit length $63$ often gives low energy error solutions when $k = 64$, the energy errors for small $k$, such as $k = 1$ and $k = 8$, are widely distributed.
Since a larger $k$ causes a longer computation time for the training of FM, one-hot FMA is superior to domain-wall FMA.
One-hot FMA often yields low energy error solutions even for small $k$.
Consequently, one-hot FMA is the best among the proposed methods from the viewpoint of the energy error distribution.

\subsubsection{Tequired number of invocations of the black-box function}
\label{sec:result-blackbox-calls-n2}
\begin{figure*}[!tb]
    \centering
    \includegraphics[width=6.8truein]{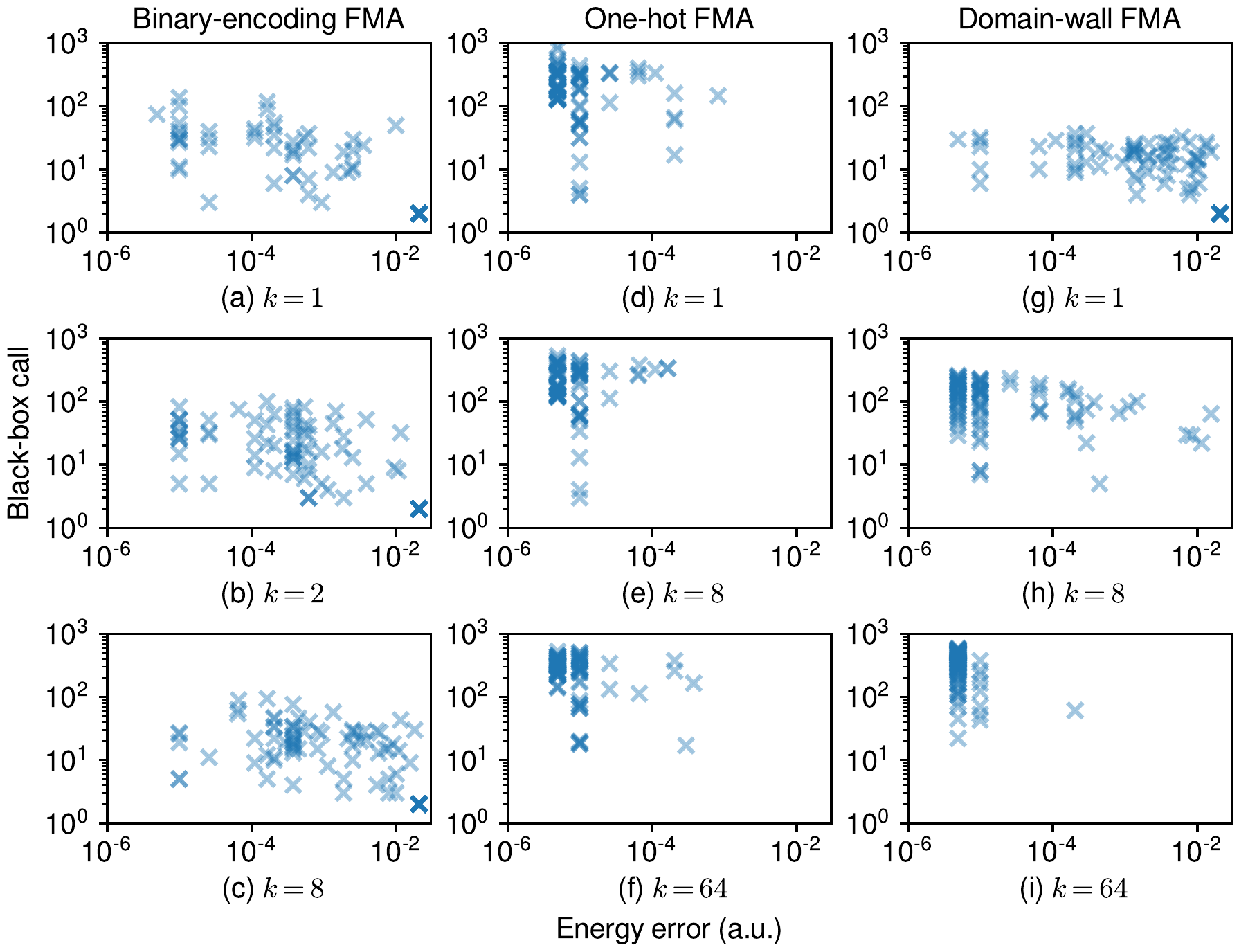}
    \caption{
        Scatterplot of the required number of invocations of the black-box function to achieve solutions with energy errors in atomic unit on the horizontal axis.
        (a)--(c) Results for binary-encoding FMA with a bit length of $6$.
        Ranks of FM are given as (a) $k=1$, (b) $k=2$, and (c) $k=8$.
        (d)--(f) Results for one-hot FMA with a bit length of $64$ and the penalty coefficient $p=1,000$.
        Ranks of FM are given as (d) $k=1$, (e) $k=8$, and (f) $k=64$.
        (g)--(i) Results for domain-wall FMA with a bit length of $63$ and the penalty coefficient $p=1,000$.
        Ranks of FM are given as (g) $k=1$, (h) $k=8$, and (i) $k=64$.
    }
    \label{fig:H2_bbc_acc_n2}
\end{figure*}

In general, the invocation of a black-box function is a bottleneck of methods for black-box optimization.
Thus, the required number of invocations is a crucial indicator to evaluate the performance.
Fig.~\ref{fig:H2_bbc_acc_n2} shows the required number of invocations of the black-box function to achieve solutions with energy errors indicated by the horizontal axis.
The value of the energy error is the least energy error in the samples obtained in a single run of FMA.
The corresponding number of invocations is the required number until FMA reaches the energy error for the first time in the run.
The data used in the scatterplot are the same as those used to plot the energy error distribution in Sec.~\ref{sec:result-energy-error-dist-n2}.

Binary encoding is not suitable for FMA.
The energy errors do not converge to a low value for the values of $k$ studied (Fig.~\ref{fig:H2_bbc_acc_n2}).
Additionally, the error does not decrease as $k$ increases.
Thus, another encoding may be more appropriate for FMA.
Both one-hot FMA and domain-wall FMA have comparable performances in terms of the required number of invocations of the black-box function.
Except for the results of domain-wall FMA with $k = 1$, both methods provide an energy error distribution that converges to a low energy error regime.
In both methods, the required number of invocations to provide a low energy error solution is on the order of $\mathrm{O}(10\text{--}10^{2})$.
The results are not significant in terms of the required number of invocations.

\subsection{Effects of increasing the search-space dimension}
\label{sec:result-increasing-dimension}

\begin{figure*}[!tb]
    \centering
    \includegraphics[width=6.8truein]{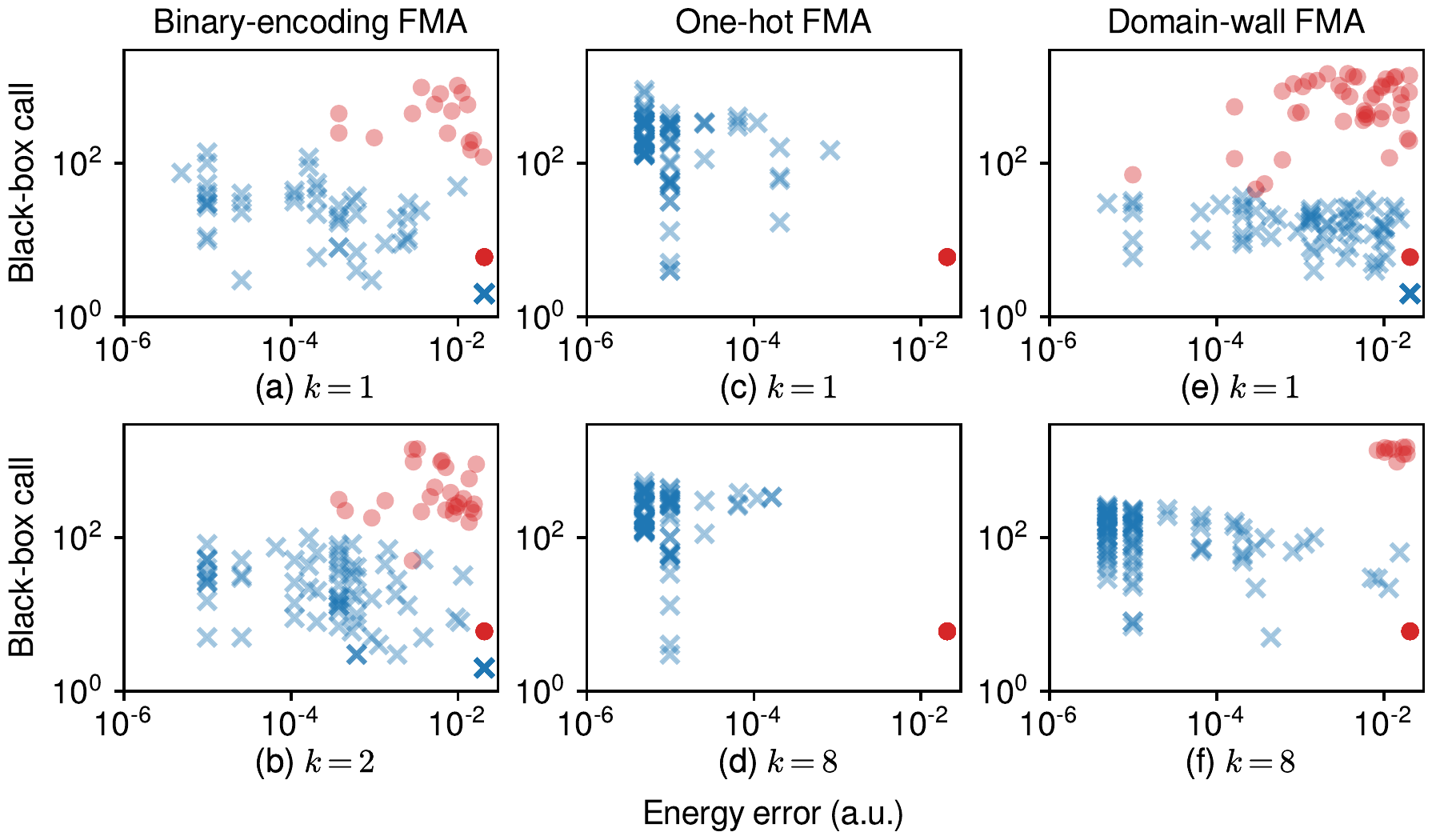}
    \caption{
        Search-space dimension dependency of the FMA performance.
        Horizontal axis represents the minimum energy error.
        Vertical axis shows the number of the invocation of the black-box function
        before the minimum energy error is achieved.
        Blue crosses and red circles indicate the performance of FMA with two-dimensional and six-dimensional search spaces, respectively.
        Rank of FM is $k=1$ for (a), (c), and (e),
        and $k=2$ for (b), $k=8$ for (d) and (f).
    }
    \label{fig:H2_bbc_acc_n2-6}
\end{figure*}

Fig.~\ref{fig:H2_bbc_acc_n2-6} shows the dependency of the FMA performance on the search-space dimensions.
In addition to the two-dimensional problem discussed in Secs.~\ref{sec:result-n2-gs-energy}--\ref{sec:result-blackbox-calls-n2}, we applied our approach to the problem with a six-dimensional search space.
Both one-hot FMA and domain-wall FMA use a penalty coefficient of $p = 1,000$.
The annealing schedule is the same as that used in the two-dimensional search space.
To obtain the distribution of the FMA performance, the two- and six-dimensional search-space problems use $100$ and $50$ runs, respectively, while changing the initial condition of FMA.

Figs.~\ref{fig:H2_bbc_acc_n2-6}(a) and \ref{fig:H2_bbc_acc_n2-6}(b) show that increasing the search-space dimensions decreases binary-encoding FMA.
The ranks of FM are (a) $k = 1$ and (b) $k = 2$.
We use six-bit integers, each of which takes an integer from $-32$ to $31$.
Both the energy error and the number of invocations of the black-box function tend to increase as the search space is expanded, indicating a degraded FMA performance.
This result is reasonable because the larger the search space, the more difficult it is to find a solution.

The results for one-hot FMA are shown in Figs.~\ref{fig:H2_bbc_acc_n2-6}(c) and \ref{fig:H2_bbc_acc_n2-6}(d).
The ranks of FM are (c) $k = 1$ and (d) $k = 8$.
We use a bit length of $d = 64$, where each integer-valued variable ranges from $-32$ to $31$.
The results for the six-dimensional search space are localized at a single point, where the energy error is equal to that of the Hartree–Fock solution.
This means that the solution using one-hot FMA is not any better than that using the Hartree–Fock solution, which is included in the initial samples.

Figs.~\ref{fig:H2_bbc_acc_n2-6}(e) and \ref{fig:H2_bbc_acc_n2-6}(f) show the results for domain-wall FMA.
The ranks of FM are (e) $k = 1$ and (f) $k = 8$.
The bit length is $d = 63$, where the variable takes the same range of integers as that for the other FMAs mentioned above.
SA generates sufficient feasible samples to train FM with a penalty coefficient $p = 1,000$.
Similar to the case of binary-encoding FMA, the performance degrades as the dimensions increase.
In particular, the required number of invocations of the black-box function increases while the energy error decreases only slightly.

In summary, expanding the search space negatively impacts all methods.
The energy error of one-hot FMA increases remarkably.
Hence, reducing the search space is crucial for the performance of the proposed FMAs.

\section{Conclusion}
\label{sec:conclusion}
This paper proposes an approach for integer-variable black-box optimization problems based on FMA, which uses Ising machines in combination with machine learning.
The performance of the proposed approach with different integer encodings was evaluated using a simple benchmark problem of calculating the ground-state energy of the hydrogen molecule.
The proposed method with all integer encodings can calculate the ground-state energy with a high accuracy.
In addition, one-hot encoding is a useful method for FMA.
However, the performance degrades as the search-space dimensions increase.
Thus, identifying the origin of performance degradation and improving the performance for larger search-space dimensions are important future works.

Since this paper uses a simple problem setting, this work can be expanded.
An example is the development of binary-encoding FMA with a restricted integer range, where some $2d$ integers must be excluded.
It is unknown whether FM can capture the properties of the low-energy state of the given problems or without a restriction.
Another example is to elucidate the relationship of the performance to training dataset for FM.
Another interesting question is whether a training dataset using only low-energy samples improves the performance.

% Appendixes, if needed, appear before the acknowledgment.
\appendices

\section{Suitable annealing schedule}
\label{sec:annealing-schedule}
The annealing schedule of SA must be adjusted to obtain low-cost samples of the given QUBO model.
The thermal fluctuations are dominant because the initial inverse temperature must be sufficiently small compared to the energy scale of the model.
However, the inverse temperature must be large so that the state converges to a local minimum at the end of the SA procedure.
Here, we discuss how to determine the initial inverse temperature for each integer encoding method.

The upper bound of the energy difference can be derived when a single bit flips.
The Hamiltonian can be denoted as
\begin{align}
    H = H_\text{FM} + p C,
\end{align}
where
\begin{align}
    C = \begin{cases}
        0 & \text{Binary encoding}, \\
        C_{\text{oh}} & \text{One-hot encoding}, \\
        C_{\text{dw}} & \text{Domain-wall encoding}.
    \end{cases}
\end{align}
The absolute value of the energy difference when flipping $x_{i}^{(l)}$ is evaluated as
\begin{align}
    \lvert \Delta H_{i}^{(l)} \rvert
    &\le
    \left\lvert \sum_{m, j} Q_{i,j}^{l,m}x_{j}^{(m)}\right\rvert + p \lvert \Delta C \rvert \notag \\
    &\le
    Ld + p \lvert \Delta C \rvert.
\end{align}
Here, the fact that $Q$ is normalized and $\lvert x_{i}^{(l)}\rvert \le 1$ for all $i$ and $l$ are used.
In the case of one-hot encoding, the constraint term is upper bounded as
\begin{align}
    \lvert \Delta C \rvert
    &\le
    \left\vert \Delta \left(\sum_{j=0}^{d-1}x_{j}^{(l)} - 1\right)^{2} \right\rvert \notag\\
    &\le
    \left\{ (d-1)^{2} - (d-2)^{2}\right\} \notag\\
    &\le
    2d-3.
\end{align}
Here, assume that $d \ge 2$.
In the case of domain-wall encoding,
\begin{align}
    \lvert \Delta C \rvert
    &\le
    2\left\vert
    \Delta \left(\sum_{j=1}^{d-1}x_{j}^{(l)} - \sum_{j=0}^{d-2}x_{j}^{(l)}x_{j+1}^{(l)} \right)
    \right\rvert
    \le
    2.
\end{align}
Therefore, the upper bound of the absolute value of the energy difference is
\begin{align}
    \Delta H_\text{upper} = \begin{cases}
        Ld & \text{Binary encoding}, \\
        Ld  + p (2d-3) & \text{One-hot encoding}, \\
        Ld + 2p & \text{Domain-wall encoding}.
    \end{cases}
\end{align}
To introduce sufficiently large thermal fluctuations at the beginning of the SA procedure, an initial inverse temperature lower than $1/\Delta H_{\text{upper}}$ is used.
The final inverse temperature cannot be determined from analytical calculations since the true ground-state energy and first-excited energy is generally unknown in advance.
Thus, preliminary simulations are necessary to determine the final inverse temperature, which leads to better results.

\section*{Acknowledgment}

This work was supported in part by the Cross-ministerial Strategic Innovation Promotion Program (SIP)
(Photonics and Quantum Technology for Society 5.0).
The computations in this work were performed using the facilities of the Supercomputer Center,
the Institute for Solid State Physics, The University of Tokyo.

\bibliography{references}

\begin{IEEEbiography}[{\includegraphics[width=1in,height=1.25in,clip,keepaspectratio]{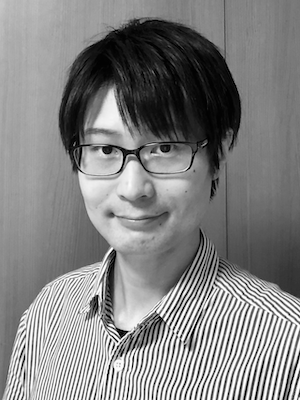}}]{Yuya Seki}
    Yuya Seki received the B.Sc., M.Sc.\ and Dr.Sc.\ degrees from the Tokyo Institute of Technology, in 2011, 2013, and 2016, respectively.
    He is currently a Project Lecturer with the Graduate School of Science and Technology,
    Keio University.
    His research interests include quantum computing, machine learning, statistical physics,
    quantum annealing, and Ising machines.
    He is a member of JPS.
\end{IEEEbiography}

\begin{IEEEbiography}[{\includegraphics[width=1in,height=1.25in,clip,keepaspectratio]{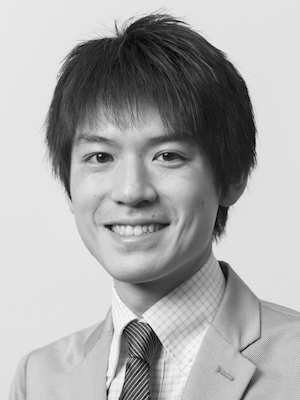}}]{Ryo Tamura}
    Ryo Tamura received the B.Sc.\ degree from the Saitama University, in 2007, and the M.Sc.\ and Dr.Sc.\ degrees from The University of Tokyo, in 2009 and 2012, respectively. He is currently a Principal Researcher with the International Center for Materials Nanoarchitectonics, National Institute for Materials Science and a Lecturer, Graduate School of Frontier Sciences, The University of Tokyo. His research interests include materials informatics, machine learning, Ising machine, and statistical physics. He is a member of the JPS.
\end{IEEEbiography}

\begin{IEEEbiography}[{\includegraphics[width=1in,height=1.25in,clip,keepaspectratio]{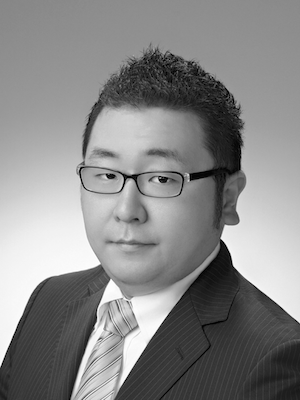}}]{Shu Tanaka}
    Shu Tanaka received the B.Sc.\ degree from the Tokyo Institute of Technology, in 2003, and the M.Sc.\ and Dr.Sc.\ degrees from The University of Tokyo, in 2005 and 2008, respectively. He is currently an Associate Professor with the Department of Applied Physics and Physico-Informatics, Keio University. His research interests include quantum annealing, Ising machine, statistical mechanics, and materials science. He is a member of the JPS.
\end{IEEEbiography}

\EOD

\end{document}